\documentclass[runningheads]{llncs}
\usepackage{graphicx}
\usepackage{xargs}
\usepackage{subfigure}
\usepackage{amsmath}
\usepackage{color,soul}
\usepackage[colorinlistoftodos,prependcaption,textsize=tiny]{todonotes}

\newcommandx{\change}[2][1=]{\todo[linecolor=blue,backgroundcolor=blue!25,bordercolor=blue,#1]{#2}}

\begin{document}
\title{Computed Tomography Image Enhancement using 3D Convolutional Neural Network}

\titlerunning{3DECNN Computed Tomography Image Enhancement}
%

\author{Meng Li\inst{1,4}\and Shiwen Shen\inst{2}\and Wen Gao\inst{1} \and William Hsu\inst{2}\and Jason Cong\inst{3,4}}

\authorrunning{Meng Li et al.}

\institute{National Engineering Laboratory for Video Technology, Peking University, China \and Department of Radiological Sciences, University of California, Los Angeles, USA \and Computer Science Department, University of California, Los Angeles, USA  
\and UCLA/PKU Joint Research Institute in Science and Engineering \\
\email{mmli@pku.edu.cn}}

\maketitle              
%
\begin{abstract}
Computed tomography (CT) is increasingly being used for cancer screening, such as early detection of lung cancer. However, CT studies have varying pixel spacing due to differences in acquisition parameters. Thick slice CTs have lower resolution, hindering tasks such as nodule characterization during computer-aided detection due to partial volume effect. In this study, we propose a novel 3D enhancement convolutional neural network (3DECNN) to improve the spatial resolution of CT studies that were acquired using lower resolution/slice thicknesses to higher resolutions.  Using a subset of the LIDC dataset consisting of 20,672 CT slices from 100 scans, we simulated lower resolution/thick section scans then attempted to reconstruct the original images using our 3DECNN network. A significant improvement in PSNR (29.3087dB vs. 28.8769dB, p-value $< 2.2e-16$) and SSIM (0.8529dB vs. 0.8449dB, p-value $< 2.2e-16$) compared to other state-of-art deep learning methods is observed. 



\end{abstract}
\section{Introduction}


Computed tomography (CT) is a widely used screening and diagnostic tool that provides detailed anatomical information on patients. Its ability to resolve small objects, such as nodules that are 1-30 mm in size, makes the modality indispensable in performing tasks such as lung cancer screening and colonography. However, the variation in image resolution of CT screening due to differences in radiation dose and slice thickness hinders the radiologist's ability to discern subtle suspicious findings. Thus, it is highly desirable to develop an approach that enhances lower resolution CT scans by increasing the detail and sharpness of borders to mimic higher resolution acquisitions~\cite{SRmedical}. 



Super-resolution~(SR) is a class of techniques that increase the resolution of an imaging system~\cite{SRoverview} and has been widely applied on natural images and is increasingly being explored in medical imaging. Traditional SR methods use linear or non-linear functions (e.g., bilinear/bicubic interpolation and example-based methods~\cite{sparse,sparse1}) to estimate and simulate image distributions. These methods, however, produce blurring and jagged edges in images, which introduce artifacts and may negatively impact the ability of computer-aided detection (CAD) systems to detect subtle nodules. Recently, deep learning, especially convolutional neural networks (CNN), has been shown to extract high-dimensional and non-linear information from images that results in a much improved super-resolution output. One example is the super-resolution convolutional neural network (SRCNN)~\cite{SRCNN}. SRCNN learns an end-to-end mapping from low- to high-resolution images. In~\cite{app1,app2}, the authors applied and evaluated the SRCNN method to improve the image quality of magnified images in chest radiographs and CT images. Moreover,~\cite{ESPCN} introduced an efficient sub-pixel convolution network (ESPCN), which was shown to be more computationally efficient than SRCNN. In~\cite{miccai2017}, the authors proposed a SR method that utilizes a generative adversarial network (GAN), resulting in images have better perceptual quality compared to SRCNN. 
All these methods were evaluated using 2D images. However, for medical imaging modalities that are volumetric, such as CT, a 2D convolution ignores the correlation between slices. We propose a 3DECNN architecture, which executes a series of 3D convolutions on the volumetric data. We measure performance using two image quality metrics: peak signal-to-noise ratio (PSNR) and structural similarity (SSIM). Our approach achieves significant improvement compared with improved SRCNN approach~(FSRCNN)~\cite{FSRCNN} and~\cite{ESPCN} on both metrics.

\section{Method}
\subsection{Overview}

For each slice in the CT volume, our task is to generate a high-resolution image $I^{HR}$ from a low-resolution image $I^{LR}$.
Our approach can be divided into two phases: model training and inference. In the model training phase, we first downsample a given image \emph{I} to obtain the low-resolution image \emph{$I^{LR}$}. We then use the original data as the high-resolution images $I^{HR}$ to train our proposed 3DECNN network. In the model inference phase, we use a previously unseen low-resolution CT volume as input to the trained 3DECNN model and generate a super resolution image $I^{SR}$. 

\begin{figure}[t]
\centering
\includegraphics[height=5.8cm]{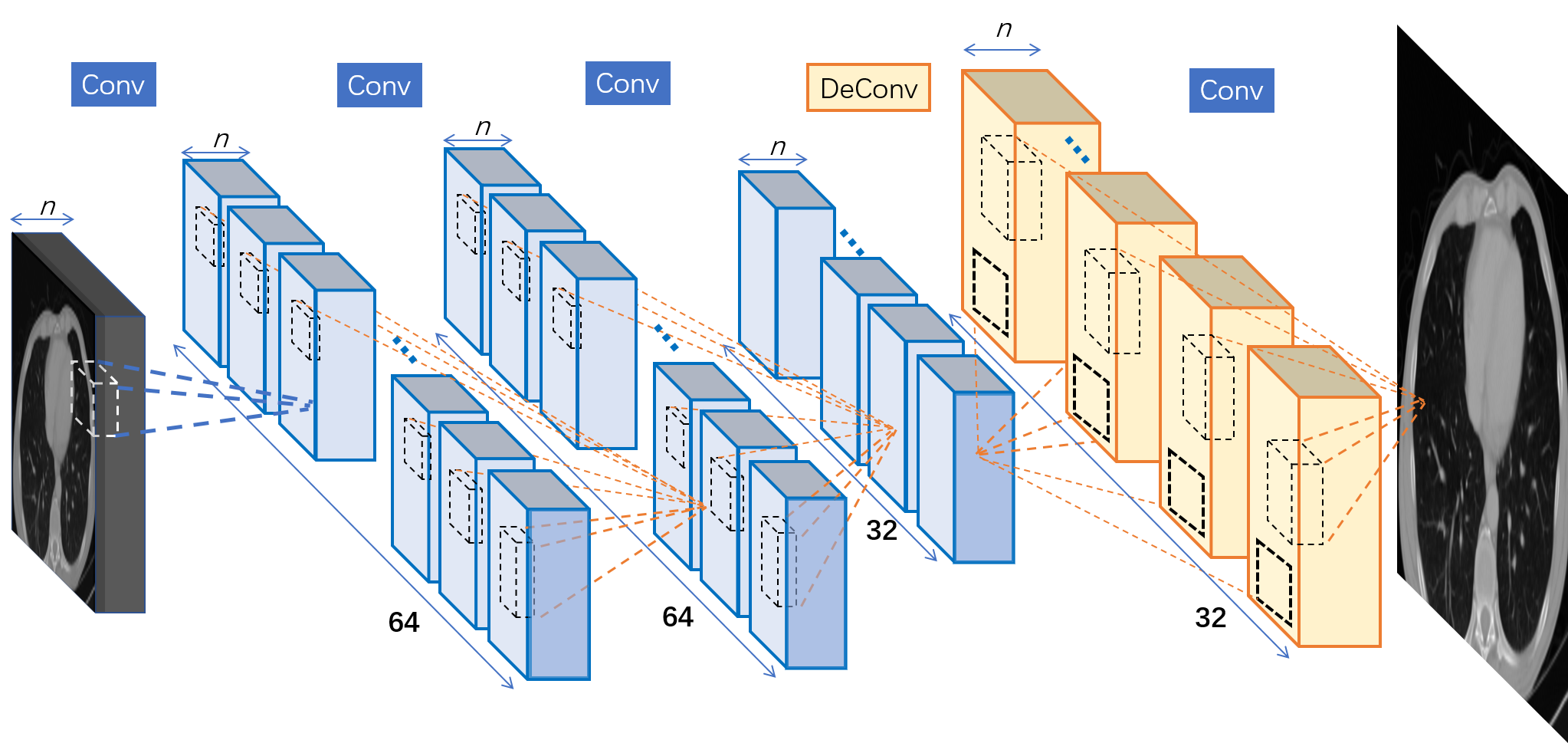}
\caption{Proposed 3DECNN architecture}
\label{fig:fig1}
\end{figure}

\subsection{Formulation}
For CT images, spatial correlations exist across three dimensions. As such, the key to generating high-quality SR images is to make full use of available information along all dimensions. Thus, we apply cube-shaped filters on the input CT slices and slides these filters through all three dimensions of the input. Our model architecture is illustrated in Fig. 1. This filtering procedure is repeated in 3 stacked layers. After the 3D filtering process, a 3D deconvolution is used to reconstruct images and up-sample them to larger ones.
The output of this 3D deconvolution is a reconstructed SR 3D volume. However, to compare with other SR methods such as SRCNN and ESPCN, which produces 2D outputs, we transform our 3D volume into a 2D output. As such, we add a final convolution layer to smooth pixels into a 2D slice, which is then compared to the outputs of the other methods. In the following paragraphs, we describe mathematical details of our 3DECNN architecture.



\subsubsection{3D Convolutional Layers.} 
In this work, we incorporate the feature extraction optimizations into the training/learning procedure of convolution kernels. The original CT images are normalized to values between [0,1]. The first CNN layer takes a normalized CT image 
(represented as a 3-D tensor) as input and generates multiple 3-D tensors (feature maps) as output by sliding the cube-shaped filters (convolution kernels), which are sized of '$k_1\times k_2\times k_3$', 
across inputs. We define convolution input tensor notations as $\langle N,C_{in},H,W \rangle$ and output $\langle N,C_{out},H,W \rangle$, in which $C_i$ stands for the number of 3-D tensors and  $\langle N,H,W \rangle$ stands for the feature map block's thickness, height, and width, respectively. Subsequent convolution layers take the previous layer's output feature maps as input, which are in a 4-D tensor. Convolution kernels are in a dimension of $\langle C_{in}, C_{out},k_1,k_2,k_3 \rangle$. The sliding stride parameter $\langle s \rangle$ defines how many pixels to skip between each adjacent convolution on input feature maps. Its mathematical expression is written as follows: $ out[c_{o}][n][h][w]=\sum_{n=0}^{C_{i}}\sum_{i=0}^{k_1}\sum_{j=0}^{k_2}\sum_{k=0}^{k_3}W[c_{o}][c_{i}][i][j][k] *In[c_{i}][{s}*n+i][{s}*h+j][{s}*w+k]$.

\subsubsection{Deconvolution layer.} In traditional image processing, a reverse feature extraction procedure is typically used to reconstruct images. Specifically, design functions such as linear interpolation, are used to up-scale images and also average overlapped output patches to generate the final SR image. In this work, we utilize deconvolution to achieve image up-sampling and reconstruct feature information from previous layers' outputs at the same time. Deconvolution can be thought of as a transposed convolution. Deconvolution operations up-sample input feature maps by multiplying each pixel with 
cubic filters and summing up overlap outputs of adjacent filters' output~\cite{deconv}. Following the above convolution's mathematic notations, deconvolution is written as the following: $ out[c_{o}][n][h[w]=\sum_{n=0}^{C_{i}}\sum_{i=0}^{k_1}\sum_{j=0}^{k_2}\sum_{k=0}^{k_3}W[c_{o}][c_{i}][i][j][k]*In[c_{i}][\frac{n}{s}+k_1-i][\frac{h}{s}+k_2-j][\frac{w}{s}+k_3-k]$.
Activation functions are used to apply an element-wise non-linear transformation on the convolution or deconvolution output tensors. In this work, we use ReLU as the activation function.

\subsubsection{Hyperparameters.}
There are four hyperparameters that have an influence on model performance: number of feature layers, feature map depth, number of convolution kernels, and size of kernels. The number of feature extraction layers $\langle l \rangle$ determines the upper-bound complexity in features that the CNN can learn from images. The feature map depth $\langle n \rangle$ is the number of CT slices that are taken in together to generate one SR image. 
The number of convolution kernels $\langle f \rangle$ decides the number of total feature maps in a layer and thus decides the maximum information that can be represented in the output of this layer. The size of convolution and deconvolution kernels $\langle k \rangle$ decides the visible scope that the filter can see in the input CT image or feature maps. 
Given the impact of each hyperparameter, we performed a grid search of the hyperparameter space to find the best combination of $\langle n,~l,~f,~k \rangle$ for our 3DECNN model.


\subsubsection{Loss function.}
Peak signal-to-noise ratio (PSNR) is the most commonly used metric to measure the quality of reconstructed lossy images in all kinds of imaging systems. A higher PSNR generally indicates a higher quality of the reconstruction image. 
PSNR is defined as the log on the division of the max pixel value over mean squared root. Therefore, we directly use the squared mean error function as our loss function: $J(w,b) = \frac{1}{m}\sum_{i=1}^{m}L(\hat{y}^{(i)},y^{(i)}) = \frac{1}{m}\sum_{i=1}^{m}||\hat{y}^{(i)}-y^{(i)}||^2$, where $w$ and $b$ represent \emph{weight} parameters and \emph{bias} parameters. $m$ is the number of training samples. $\hat{y}$ and $y$ refer to the output of the neural network and the target, respectively. In addition, the target loss function is minimized using stochastic gradient descent with the back-propagation algorithm~\cite{backpro}. 

\section{Experiments and Results}

In this section, we first introduce the experiment setup, including dataset and data preparation. Then we show the design space of the hyper-parameters, at which time we show how to explore different CNN architectures and find the best model. Subsequently, we compare our method with recent state-of-the-art work and demonstrate the performance improvement. Lastly, we present examples of the generated SR CT images using our proposed method and previous state-of-the-art results. 
\begin{figure}[t] 
  \centering 
  \subfigure[Influence of feature map depth]{ 
    \label{fig:subfig:a} 
    \includegraphics[width=0.48\linewidth]{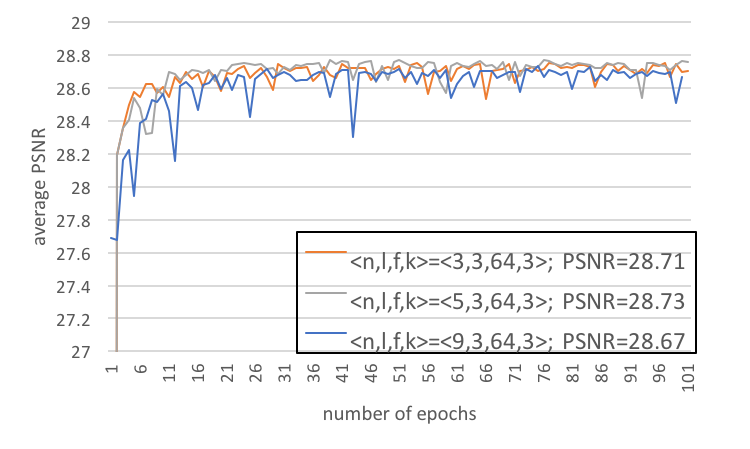}} 
  \subfigure[Infulence of the number of layers]{ 
    \label{fig:subfig:b} 
    \includegraphics[width=0.48\linewidth]{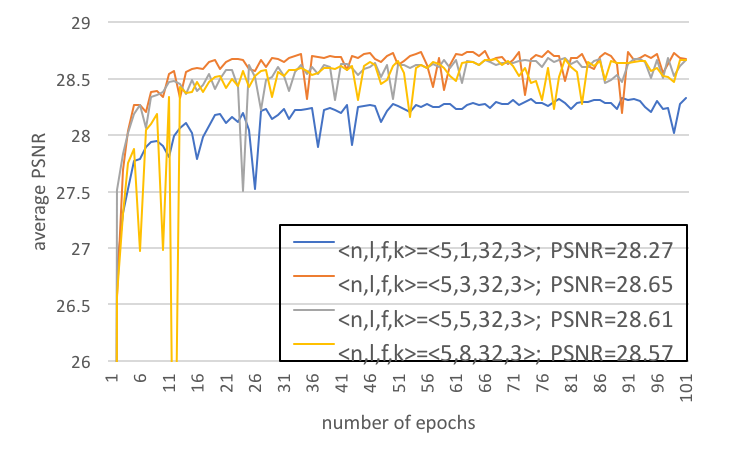}} 
  \subfigure[Influence of the number of kernels]{ 
    \label{fig:subfig:c} 
    \includegraphics[width=0.48\linewidth]{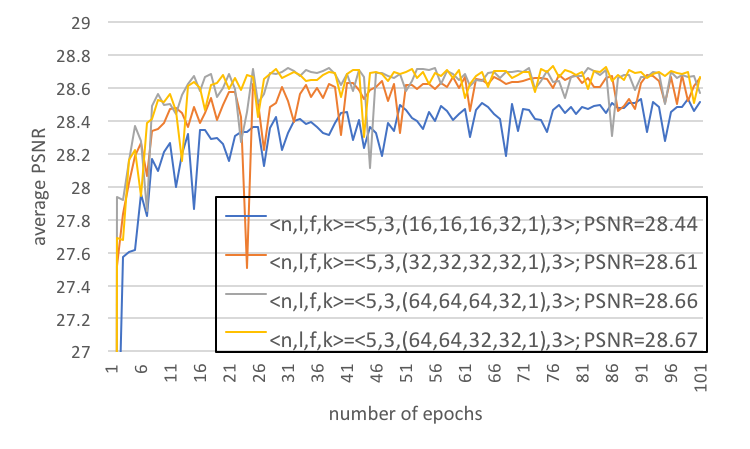}} 
  \subfigure[Influence of convolutional kernel size]{ 
    \label{fig:subfig:d} 
    \includegraphics[width=0.48\linewidth]{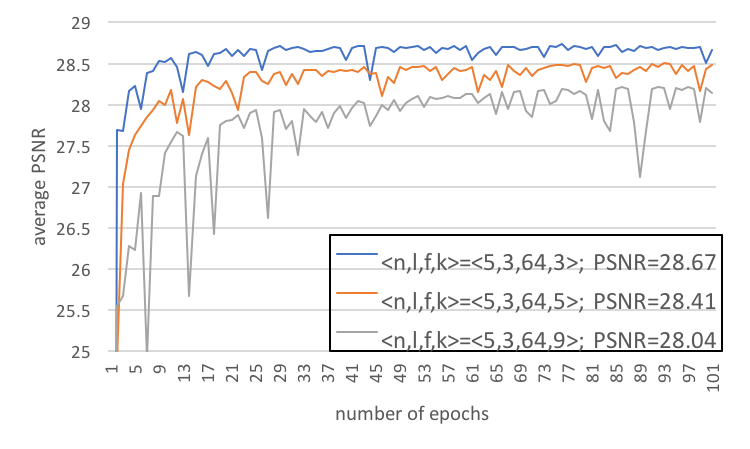}} 
  \caption{Design space of hyper-parameters} 
  \label{fig:1} 
\end{figure}


\subsection{Experiment setup}
\subsubsection{Dataset.}
We use the public available Lung Image Database Consortium image collection (LIDC) dataset for this study~\cite{LIDC_2011}, which consists of low- and diagnostic-dose thoracic CT scans. These scans have a wide range of slice thickness ranging from $0.6$ to $5$ $mm$. And the pixel spacing in axial view (x-y direction) ranges from $0.4609$ to $0.9766$ $mm$. We randomly select 100 scans out of a total of 1018 cases from the LIDC dataset, result in a total consisting of 20672 slices. The selected CT scans are then randomized into four folds with similar size. Two folds are used for training, and the remaining two folds are used for validation and test, respectively.

\subsubsection{Data preprocessing.}
For each CT scan, we first downsample it on axial view by the desired scaling factor (set 3 in our experiment) to form the LR images. Then the corresponding HR images are ground truth images.

\subsubsection{Hyperparameter tuning $\langle n,~l,~f,~k \rangle$.}
We choose the four most influential parameters to explore in our experiment and discuss, which is \emph{feature depth (n)}, \emph{number of layers (l)}, \emph{number of filters (f)} and \emph{filter kernel size (k)}.

The effect of the \textbf{feature depth $\langle n\rangle$} is shown in Fig.~\ref{fig:subfig:a}. It presents the training curves of three different 3DECNN architectures, in which their $\langle l,~f,~s \rangle$ are the same and $\langle n \rangle$ varies in $[3,~5,~9]$. Among the three configurations, $n=3$ has a better average PSNR than the others.
The effect of the \textbf{number of layers $\langle l \rangle$} is shown Fig.~\ref{fig:subfig:b}, which demonstrates that a deeper CNN may not always be better. With fixed $\langle n,~f,~s \rangle$ and varying $l \in [1,3,5,8]$, here $l$ indicate the number of convolutional layers before the deconvolution process. we can observe apparent different performance on the training curves. We determine that $l=3$ achieves higher average PSNR.
The effect of the \textbf{number of filters $\langle f \rangle$} is shown in Fig.~\ref{fig:subfig:c}, in which we fix $\langle n,~l,~k \rangle$ and choose $\langle f \rangle$ in four collections.
An apparent drop in PSNR is seen when $\langle f \rangle$ chooses the too small configuration $\langle 16,~16,~16,~32,~1 \rangle$. $\langle 64,~64,~64,~32,~1 \rangle$ and $\langle 64,~64,~32,~32,~1 \rangle$ has approximately the same PSNR (28.66 vs. 28.67) so we choose latter one to save training time. 
The effect of the \textbf{filter kernel size $\langle k \rangle$} is shown in Fig.~\ref{fig:subfig:d}, in which we fix $\langle n,~l,~f \rangle$ and vary k in the collection of $[3,~5,~9]$. Experiment result proves that $k=3$ achieves the best PSNR. The PSNR decrease with filter kernel size demonstrate that relatively remote pixels contribute less to feature extraction and bring much signal noise to the final result.




\subsubsection{Final model.}
For the final design, we set $\langle~ n, l, (f_1,k_1),(f_2,k_2),(f_3,k_3),(f_4^{deconv},\\k_4^{deconv}),(f_5,k_5) ~\rangle$ 
$ = \langle 5, 3,(64,3),(64,3), (32,3),(32,3),(1,3) \rangle$. We set the learning rate $\alpha$ as $10^{-3}$ for this design and achieve a good convergence. We implemented our 3DECNN model using Pytorch and trained/validated our model on a workstation with a NVIDIA Tesla K40 GPU. The training process took roughly 10 hours. 
\begin{table}[t]
\centering
\caption{PSNR and SSIM results comparison.}
\includegraphics[height=3cm]{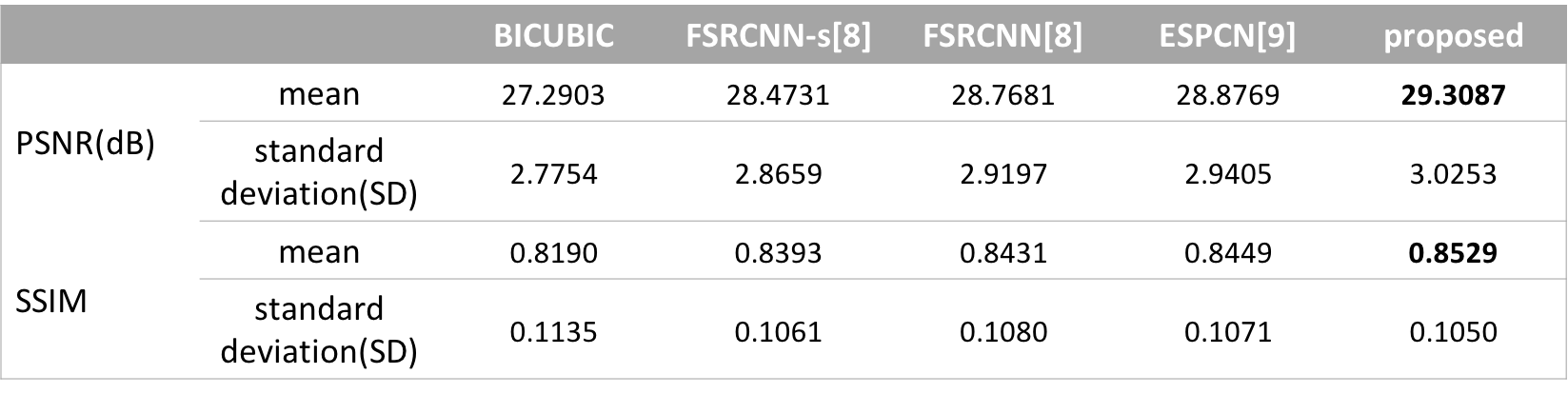}
\label{table:table1}
\end{table}





\subsection{Results comparison with T-test validation}
\begin{figure}[t]
\centering
\includegraphics[height=8cm]{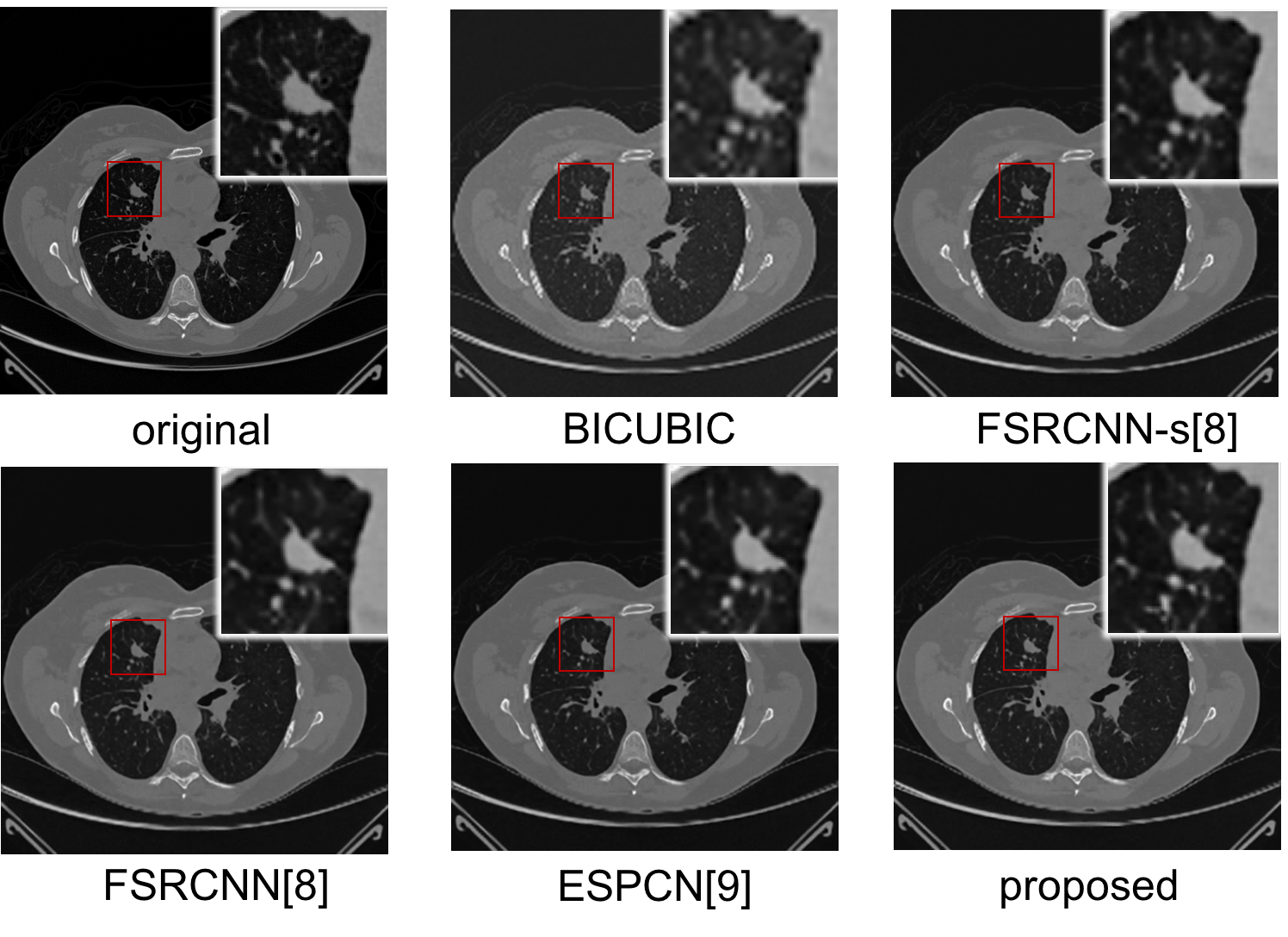}
\caption{Comparison with the-state-of-the-art works}
\label{fig:fig3}
\end{figure}

We compare the proposed model to bicubic interpolation and two existing the-state-of-the-art deep learning methods for super resolution image enhancement: 1) FSRCNN~\cite{FSRCNN} and 2) ESPCN~\cite{ESPCN}. We reimplemented both methods, retraining and testing them in the manner as our proposed method. Both the FSRCNN-s and the FSRCNN architectures used in~\cite{FSRCNN} are compared here. A paired t-test is adopted to determine whether a statistically significant difference exists in mean measurements of PSNR and SSIM when comparing 3DECNN to bicubic, FSRCNN, and ESPCN. Table 1 shows the mean and standard deviation for the four methods in PSNR and SSIM using 5,168 test slices. The paired t-test results show that the proposed method has significantly higher mean PSNR, and mean differences are 2.0183 dB (p-value $< 2.2e-16$), 0.8357 dB (p-value $< 2.2e-16$), 0.5406 dB (p-value $< 2.2e-16$), and 0.4318 dB (p-value $< 2.2e-16$) for bicubic, FSRCNN-s, FSRCNN and ESPCN, respectively. 
It also shows that out model has significantly higher SSIM, and the mean differences are $0.0389$ (p-value $< 2.2e-16$), $0.0136$ (p-value $< 2.2e-16$), $0.0098$ (p-value $< 2.2e-16$), and $0.0080$ (p-value $< 2.2e-16$). 
To subjectively measure the image perceived quality, we also visualize and compare the enhanced images in Fig. \ref{fig:fig3}. The zoomed areas in the figure are lung nodules. As the figures shown, our approach achieved better perceived quality compared to other methods.

\section{Discussion and Future work}
We present the results of our proposed 3DECNN approach to improve the image quality of CT studies that are acquired at varying, lower resolutions. Our method achieves a significant improvement compared to existing state-of-art deep learning methods in PSNR (mean improvement of $0.43 dB$ and p-value $< 2.2e-16$) and SSIM (mean improvement of $0.008$ and p-value $< 2.2e-16$). We demonstrate our proposed work by enhancing large slice thickness scans, which can be potentially applied to clinical auxiliary diagnosis of lung cancer. As future work, we explore how our approach can be extended to perform image normalization and enhancement of ultra low-dose CT images (studies that are acquired at 25\% or 50\% dose compared to current low-dose images) with the goal of producing comparable image quality while reducing radiation exposure to patients.

\section{Acknowledgement}
This work is partly supported by National Natural Science Foundation of China (NSFC) Grant 61520106004, the National Institutes for Health under award No. R01CA210360. The authors would also like to thank the UCLA/PKU Joint Research Institute, Chinese Scholarship Council for their support of our research.

\end{document}